\documentclass{article}

\usepackage{PRIMEarxiv}

\usepackage[utf8]{inputenc} 
\usepackage[T1]{fontenc}    
\usepackage{hyperref}       
\usepackage{url}            
\usepackage{booktabs}       
\usepackage{amsfonts}       
\usepackage{nicefrac}       
\usepackage{microtype}      
\usepackage{lipsum}
\usepackage{fancyhdr}       
\usepackage{graphicx}       
\graphicspath{{media/}}     

\title{Entropy in Large Language Models}

\author{
  Marco Scharringhausen \\
  Carl von Ossietzky Universität Oldenburg \\
  Oldenburg, Germany\\
  \texttt{marco.scharringhausen@uol.de} \\
}

\begin{document}
\maketitle

\begin{abstract}
In this study, the output of large language models (LLM) is considered an information source generating an unlimited sequence of symbols drawn from a finite alphabet. Given the probabilistic nature of modern LLMs, we assume a probabilistic model for these LLMs, following a constant random distribution and the source itself thus being stationary. 

We compare this source entropy (per word) to that of natural language (written or spoken) as represented by the Open American National Corpus (OANC). Our results indicate that the word entropy of such LLMs is lower than the word entropy of natural speech both in written or spoken form.

The long-term goal of such studies is to formalize the intuitions of \emph{information} and \emph{uncertainty} in large language training to assess the impact of training an LLM from LLM generated training data. This refers to texts from the world wide web in particular.

\end{abstract}


\section{Introduction}
\label{sec:intro}
 
Language models (often referred to as large language models, or LLMs) are machine learning ("AI") systems that are trained on vast amounts of text. These models are based on neural networks and machine learning methods. They aim at understanding, analyzing, and generating human language on themselves. LLMs recognize patterns and relationships in language. They are able to perform tasks such as text summaries, translation, chat completions, and content creation (text, code). 

Language models largely impacted the way we shape and consume the world wide web and literature in the recent past. In particular the rise of transformer neural networks \cite{vaswani2023attentionneed} provided a leap forward in terms of the linguistic quality and capabilities of such tools. Recent years' progress in this scientific and technological field is largely based on three developments: increased availability and amount of data (also: improved quality of same), growing computing power (e.g. cloud computing) and powerful algorithms (such as \cite{vaswani2023attentionneed}).

More and more content is generated using AI tools. This applies to images and program code and mostly to text generated by LLMs. At the same time, language models are trained from large reservoirs of text. Usually, the world wide wed is the natural reservoir for such a purpose. Thus, it is self-evident to raise the question of what the long-term effects will be. It is to be investigated wether training of large language models using "artificial" text (generated by other language models) might change the linguistic quality of the text that is being generated by later models.

Since this question is not easy to answer, we will approach the high-level research question using mathematical tools from elementary statistics. In written as well as in spoken language, information is encoded by a sequence of basic symbols (the "alphabet") under certain and language-specific grammatical-semantic constraints. The entropy rate per word quantifies the amount of information that is encoded in this stream of symbols. 

We assume that any LLM generates text from a stationary (i.e. not time-dependent) probability distribution either on letter or word level. In fact, it is the strength of modern LLMs and their architecture to explicitly make use of such long-range dependencies in the input data.  We will be using the concept of entropy rate to estimate the entropy of the LLM output that is considered to the the source (see section \ref{sec:sourcentropy}). Using the notion of entropy rate enables research to quantify the amount of information inside the individual large language models. 

This study aims at cross-secting through the field of LLMs to represent a broad use case. That is, the language models are accessed in a way the "common user" would do. Although we will be using an API for sake of efficiency, the input prompts are close to everyday life and the input parameters are mostly assumed to be the respective default values. The only exception will be the model temperature $T$.

\section{Methodology}

To obtain a general result, a large number of large language models is being utilized in this study. These LLMs are accessed using APIs to generate a large amount of text output. The results are fed into a Python tool calculating the entropy rate.

\subsection{LLMs in This Study}
\label{sec:llmparams}

This study includes two groups of large language models: Those from the \emph{Blablador} tool \cite{blablador2023} and those hosted by \emph{Mistral} \cite{mistral2023}. 

Blablador is an evaluation server for Large Language Models hosted in the Helmholtz community. Users are offered a chat interface as well as an API \cite{blablador2023api}. According to its self-description, Blablador is not ready for production use and rather suited for experiments. This is sufficient for our purposes in this study.

Mistral AI is a French artificial intelligence company, founded in the early 2020's. Mistral offers open-weight large language models as well as proprietary models. Beside a web ui, Mistral offers API access to their models as well \cite{mistral2023api}.

This study includes in total 18 different large language models with parameter numbers ranging from 3 up to 230, see table \ref{tab:llmtable}.

\begin{table}
 \caption{Models of this study. Number of parameters are listed in billions.}
\label{tab:llmtable}
  \centering
  \begin{tabular}{lll}
    \toprule
    Name     & \#Parameters / B & \#Active parameters / B  \\
    \midrule
    Ministral-3-14B-Instruct-2512   &  14    &        \\
    GPT-OSS-120b                    &  117   &  5.1   \\
    MiniMax-M2                      &  230   &  10    \\
    Apertus-8B-Instruct-2509        &  8     &        \\
    Qwen3 235B - A22B               &  235   &  22    \\
    Qwen3-Coder-30B-A3B-Instruct    &  30    &  3     \\
    Phi-4-multimodal-instruct       &  5.6   &        \\
    Qwen3-VL-32B-Instruct-FP8       &  32    &        \\
    Tongyi-DeepResearch-30B-A3B     &  30    &  3     \\
    Ministral-8B-Instruct-2410      &  8B    &        \\
    Mistral-large-2512              &  675   &   41   \\
    Mistral-medium-2508             &        &        \\
    Ministral-14B-2512              &  14    &        \\
    Mistral-small-2506              &  24    &        \\
    Ministral-8B-2512               &  8     &        \\
    Ministral-3B-2512               &  3     &        \\
    Magistral-small-2509            &  24    &        \\
    Magistral-medium-2509           &  24.4  &        \\
    \bottomrule
  \end{tabular}
\end{table}

\subsection{Generation of LLM output}
\label{sec:llmoutput}

The LLMs in this study are accessed via API \cite{mistralapi}, \cite{openaiapi}. This is a high-level study. A detailed study of the impact of individual LLM input parameters on the output source entropy is beyond the scope of this study. We thus keep most of the parameters as the default values in each respective API \cite{openaiapidefaults}, \cite{mistralapidefaults}. The only deviation from the default values is the choice of model temperature. We run each model and each input set (see below) for $T = 0.3$, $T = 0.5$ and $T = 0.7$.

the role of the model temperature can be summarized as follows. LLMs generate text by predicting the next word (or rather, the next token) according to a probability distribution. Each token is assigned a numerical value. The total set of tokens is normalized to a “softmax probability distribution” with values between zero and one. By normalization, the sum of all the tokens’ softmax probabilities is one.

The LLM temperature parameter modifies this distribution. A lower temperature essentially makes those tokens with the highest probability more likely to be selected; a higher temperature increases a model's likelihood of selecting less probable tokens. This is the case since larger temperature value introduces more variability into the LLM's token selection. Different temperature settings essentially introduce different levels of randomness when a generative AI model outputs text.

To gain a large amount of text, we prompt the models with a large list of input items. The first part of the input prompt is the same for every input item, i.e. "Write an essay about ..." or "Write an essay about the ..." and then an item from the following lists:

\begin{itemize}
    \item Countries of the world (201 items)
    \item Capitals of above countries (201 items)
    \item European apples (251 items)
    \item European and asian mammals (227 items)
    \item Animals in general (139 items)
    \item Physical constants (e.g. speed of light) (54 items)
\end{itemize}

These lists are processed for every of the large language models in the list above (see table \ref{tab:llmtable}). this generates an ASCII text file that is handed over to a small Python code that calculates the entropy rate (section \ref{sec:sourcentropy}).

Apart from the model temperature $T$, the individual default values for the different LLMs are used. We assume that the "standard" user does not change the default values (with the exception of temperature maybe). This is exactly the behavior we want to mimic. The results of this study shall be representative for the "common, non-specialized and non-experienced" user.

\subsection{Open American National Corpus}
\label{sec:oanc}

The Open American National Corpus (OANC) is a electronic collection of American English. This corpus is publicly available (\cite{oanc-online}, \cite{oanc-paper}) without any restrictions. It includes texts of all genres and transcripts of spoken data produced from 1990 onward.

The American National Corpus (ANC) project, on the other hand, is not publicly available. The ANC aims at collecting a corpus comparable to the British National Corpus (BNC, \cite{bnc-book}. Corpus-analytic work has demonstrated that the BNC is inappropriate for the study of American English, due to the numerous differences in use of the language. 

The OANC is a collaborative development project that relies on contributions of data and annotations from the linguistics and natural language processing communities as well as the public at large.

This study will be using the OANC since it is the only one of the three corpi that is publicly available. The OANC contains both written as well as spoken (and then trans-scripted) text. The written part contains approximately 11.5 million word, the spoken part approximately 3 million words (see \ref{tab:llmoutputsize}).

\subsection{Source entropy}
\label{sec:sourcentropy}

The \emph{Shannon entropy per letter} is likely not the best way to estimate the entropy of the LLM output as a whole. Given a finite alphabet $X = \{ x_1,\dots,x_n \}$ and a discrete probability distribution  $p_1,\dots,p_n$, the Shannon entropy per letter is given by

\begin{eqnarray}
 H(X) &=& \frac{1}{n} \sum_{i=1}^n p_i \log(p_i)
\end{eqnarray}

Assuming that individual letters or words are stochastically independent is too restrictive in many cases when describing real sources. In natural languages, there are dependencies between successive letters and words. A concept of entropy taking into account such dependencies can be realized using the \emph{entropy rate}. We consider the LLM output as a stationary source of information and estimate the LLM output entropy rate as the limit of conditional entropies with successively longer context lengths $n$:

\begin{eqnarray}
 H(X) &=& \lim_{n\to \infty} \frac{1}{n} H_n(X_n, \dots, X_1) \\
      &=& \lim_{n\to \infty} H_n(X_n | X_{n-1}, \dots, X_1) \label{eq:limitentropy}
\end{eqnarray}

We use a python code \cite{pitclaudel2016} to calculate this entropy rate for the different LLM output texts. During these calculations, we use an upper bound of $n=6$. It should be noted that even this introduces undersampling in the results, since we use text sizes of some million words (see table \ref{tab:llmoutputsize}). Assuming an alphabet of size  approximately 45 (letters, digits, special characters), a context length of 4 yields $45^4 \approx 4 \cdot 10^6$ combinations. Thus, the number of samples in a text of some million words lengths is probably low. 

Despite these restrictions, we will use context lengths of up to 6 for the estimation of the limit given in equation (\ref{eq:limitentropy}).

\section{Results}
\label{sec:results}

The different models generate different amounts of text. Depending on the model type, architecture and temperature, outputs have sizes of 7 to 26.5 million words (table \ref{tab:llmoutputsize}). The only exception are the Blablador models for the highest temperature $T=1.5$. Here, the output size is only 140.000 words. The reason for this is yet unclear and will not be investigated further in this study. The OANC (Open American National Corpus, section \ref{sec:oanc}) represents the reference of natural language to compare the entropy rates of the LLMs to.

\begin{table}
 \caption{Model output sizes of this study. Counts are number of million words in the respective data. Depending on the model and the temperature, the output size varies. Therefore, usually a range of word counts is listed. "Accumulated" sizes are concatenations of all output files for the Mistral or the Blablador model family. "LLM accumulated" contains all models and all temperatures $T = 0.3 \dots 0.7"$}
  \centering
  \begin{tabular}{lll}
    \toprule
    Model                                        & Word count / $10^6$ \\
    \midrule
    OANC spoken                                  & 3.1       \\
    OANC written                                 & 11.5      \\
    LLM accumulated, $T = 0.3 \dots T = 1.5$     & 101.5     \\
    \midrule
    LLM accumulated, $T = 0.3$                   & 24.1      \\
    LLM accumulated, $T = 0.5$                   & 24.7      \\
    LLM accumulated, $T = 0.7$                   & 26.5      \\
    LLM accumulated, $T = 1.0$                   & 15.5      \\
    LLM accumulated, $T = 1.5$                   & 10.7      \\
    Mistral accumulated, $T = 0.3$               & 15.3      \\
    Mistral accumulated, $T = 0.5$               & 15.2      \\
    Mistral accumulated, $T = 0.7$               & 16.4      \\
    Mistral accumulated, $T = 1.0$               & 8.4       \\
    Mistral accumulated, $T = 1.5$               & 10.6      \\
    Blablador accumulated, $T = 0.3$             & 8.8       \\
    Blablador accumulated, $T = 0.5$             & 9.6       \\
    Blablador accumulated, $T = 0.7$             & 10.1      \\
    Blablador accumulated, $T = 1.0$             & 7.0       \\
    Blablador accumulated, $T = 1.5$             & 0.14      \\
    \bottomrule
  \end{tabular}
  \label{tab:llmoutputsize}
\end{table}

After calculating the entropy rate for context sizes $1 \dots 6$, the data is extrapolated towards infinity using a least squares fit based on a simple exponential model:

\begin{eqnarray}
    f(x) = a \cdot e^{-b\cdot x} + c \label{eqn:fitfunc}
\end{eqnarray}

\begin{figure}
  \centering
  \includegraphics{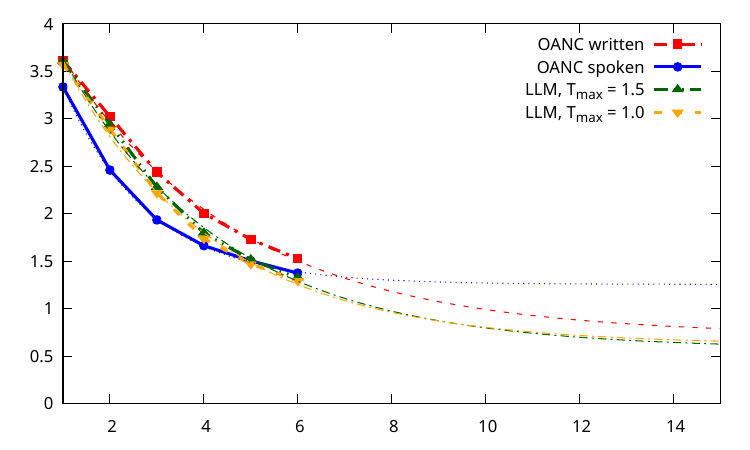}
  \caption{Entropy rates for the OANC and the output of all LLMs (both Mistral and Blablador model family concatenated) for $T=0.3\dots1.5$ as well as for $T=0.3\dots1.0$. The difference between $T_{\rm{max}}=1.0$ and $T_{\rm{max}}=1.5$ is small (0.618 and 0.574, respectively), probably since the amount of LLM output with somewhat erroneous entropy (see \ref{fig:fig2}) is small. See table \ref{tab:sourceentropy} for numeric values of the entropy rates.}
  \label{fig:fig1}
\end{figure}

\begin{figure}
  \centering
  \includegraphics{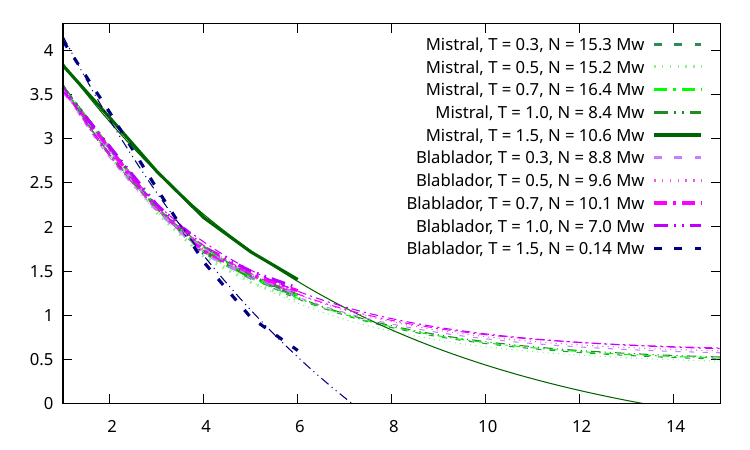}
  \caption{Entropy rates for the two model families Mistral and Blablador, distinguished by model temperature. The Blablador output for  $T=1.5$ leads to very little output (140.000 words). This leads to an estimate of the entropy rate that is most likely wrong, see \ref{sec:sourcentropy} and the comment about sample sizes therein. The impact of model temperature on the entropy rate of the output is small. The values for the Mistral model family ($0.450\dots0.475$) are lower than for the Blablador family ($0.530\dots0.587$). It is not clear, however, wether this is a significant effect.}
  \label{fig:fig2}
\end{figure}

The entropy rate as per equation \ref{eq:limitentropy} is equal to $c$. Table \ref{tab:sourceentropy} shows the values $c$. Figures \ref{fig:fig1} and \ref{fig:fig2} show the results for all LLMs (Blablador and Mistral, figure \ref{fig:fig1}) and the different LLM families for individual temperatures $T=0.3,0.5,0.7,1.0,1.5$, see figure \ref{fig:fig2}.

As can be seen from figure \ref{fig:fig2}, the temperature $T$ has little impact on the entropy rate of the individual models. This study's results do not show large differences in the outputs for temperatures between, $0.3$ and $1.5$ or $1.0$ for the Mistral and the Blablador model family, respectively. The values for the Mistral model family ($0.450\dots0.475$) are lower than for the Blablador family ($0.530\dots0.587$). It is not clear, however, wether this is a significant effect.

Note that in figure \ref{fig:fig2}, the Blablador output for $T=1.5$ as well as for Mistral and $T=1.5$ is excluded and plotted individually, since $T=1.5$ leads to very little values of the entropy. In the Blablador case, this might be a sampling issues due to very little output (140.000 words). This leads to an estimate of the entropy rate that is most likely wrong, see \ref{sec:sourcentropy} and the comment about sample sizes therein. 

Comparing the entropy rate of LLM output ($T_{\rm{max}}=1.0$ or $T_{\rm{max}}=1.5$) across all large language models, it is evident that the entropy is lower that that of natural speech (as represented by the OANC). The OANC language samples for written text constitute entropy rate of approx. 0.716 per word, whereas the entropy rate for spoken language is approximately 1.26 (see table \ref{tab:sourceentropy}). In comparison, large language models generate texts with entropy rates of approximately 0.574 and 0.618 ($T_{\rm{max}}=1.5$ and $T_{\rm{max}}=1.0$, respectively).

The root cause for lower entropy rates in LLMs compared to natural language are beyond the scope of this paper. In-depth investigation of this observation will certainly require detailed knowledge of linguistics and about the working principles of large language models, mainly pre-trained transformers.

\begin{table}
 \caption{Source entropy for different sources.}
  \centering
  \begin{tabular}{lll}
    \toprule
    Source                                       & Entropy / bit/word \\
    \midrule
    OANC written                                 & 0.716      \\
    OANC spoken                                  & 1.255      \\
    LLM accumulated, $T_{max} = 1.5$             & 0.574      \\
    LLM accumulated, $T_{max} = 1.0$             & 0.618      \\
    \midrule 
    Mistral accumulated, $T = 0.3$               & 0.450      \\
    Mistral accumulated, $T = 0.5$               & 0.433      \\
    Mistral accumulated, $T = 0.7$               & 0.463      \\
    Mistral accumulated, $T = 1.0$               & 0.475      \\
    Mistral accumulated, $T = 1.5$               & $< 0$      \\
    Blablador accumulated, $T = 0.3$             & 0.530      \\
    Blablador accumulated, $T = 0.5$             & 0.549      \\
    Blablador accumulated, $T = 0.7$             & 0.587      \\
    Blablador accumulated, $T = 1.0$             & 0.571      \\
    Blablador accumulated, $T = 1.5$             & $< 0$      \\
    \bottomrule
  \end{tabular}
  \label{tab:sourceentropy}
\end{table}

\section{Conclusion}

The entropy rates of extensive texts samples from large language models have been calculated using a small python tool. Two families of LLMs have been investigated, those from the Mistral service as well as those provided by the Blablador service. Apart from the model temperature, all parameters have been left to the default. 

A dependence on the model temperature has not been found. Entropy rates (per word) for the Mistral models are lower than that of the Blablador models, although the difference is small and it is unclear wether it is significant.

The results of all calculations indicate that large language models generate text with lower entropy rates (per word) than natural speech as represented by the Open American National Corpus (OANC). This observation is to be investigated in further detail.

The further meaning of this observation is subject of future investigation. These results indicate, however, that the statistical structure of LLM generated text is somewhat different from natural language.

\bibliographystyle{unsrt}  
\bibliography{references}  



\end{document}